\documentclass[11pt,a4paper]{article}
\usepackage[hyperref]{acl2021}
\usepackage{times}
\usepackage{latexsym}
\usepackage{booktabs}
\usepackage{multirow}
\usepackage{multicol}
\usepackage{graphicx}
\usepackage{amsmath}
\usepackage{amsfonts}
\usepackage{bbm}
\usepackage{enumitem}

\usepackage{microtype}

\aclfinalcopy 

\newcommand{\astfootnote}[1]{
    \let\oldthefootnote=\thefootnote
    \setcounter{footnote}{1}
    \renewcommand{\thefootnote}{\fnsymbol{footnote}}
    \footnotetext{#1}
    \let\thefootnote=\oldthefootnote
}

\title{Self-Guided Contrastive Learning for BERT Sentence Representations}

\author{Taeuk Kim$^{\dagger*}$, Kang Min Yoo$^{\ddagger}$, and Sang-goo Lee$^{\dagger}$ \\
$^{\dagger}$Dept. of Computer Science and Engineering, Seoul National University, Seoul, Korea \\
$^{\ddagger}$NAVER AI Lab, Seongnam, Korea \\
{\tt \{taeuk,sglee\}@europa.snu.ac.kr, kangmin.yoo@navercorp.com}}

\date{}

\begin{document}
\maketitle
\begin{abstract}
Although BERT and its variants have reshaped the NLP landscape, it still remains unclear how best to derive sentence embeddings from such pre-trained Transformers.
In this work, we propose a contrastive learning method that utilizes self-guidance for improving the quality of BERT sentence representations.
Our method fine-tunes BERT in a self-supervised fashion, does not rely on data augmentation, and enables the usual \verb|[CLS]| token embeddings to function as sentence vectors.
Moreover, we redesign the contrastive learning objective (NT-Xent) and apply it to sentence representation learning.
We demonstrate with extensive experiments that our approach is more effective than competitive baselines on diverse sentence-related tasks. 
We also show it is efficient at inference and robust to domain shifts.
\end{abstract}

\section{Introduction}

\astfootnote{This work has been mainly conducted when TK was a research intern at NAVER AI Lab.}
\setcounter{footnote}{0}

Pre-trained Transformer \cite{vaswani2017attention} language models such as BERT \cite{devlin-etal-2019-bert} and RoBERTa \cite{liu2019roberta} have been integral to achieving recent improvements in natural language understanding.
However, it is not straightforward to directly utilize these models for sentence-level tasks, as they are basically pre-trained to focus on predicting (sub)word tokens given context.
The most typical way of converting the models into sentence encoders is to fine-tune them with supervision from a downstream task.
In the process, as initially proposed by \citet{devlin-etal-2019-bert}, a pre-defined token's (a.k.a. \verb|[CLS]|) embedding from the last layer of the encoder is deemed as the representation of an input sequence.
This simple but effective method is possible because, during supervised fine-tuning, the \verb|[CLS]| embedding functions as the only communication gate between the pre-trained encoder and a task-specific layer, encouraging the \verb|[CLS]| vector to capture the holistic information.

\begin{figure}[t!]
\begin{center}
\includegraphics[width=\linewidth]{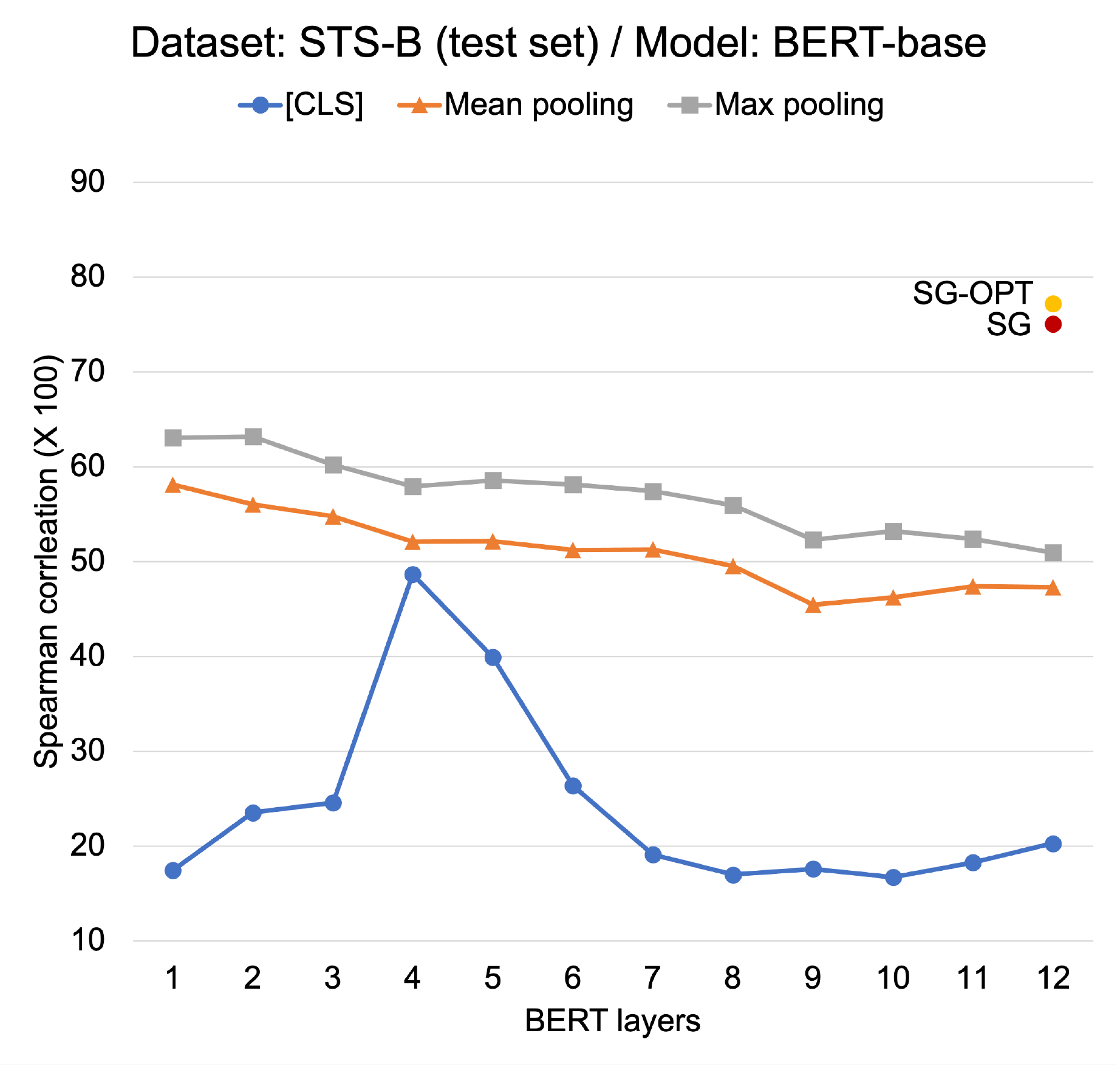}
\caption{BERT(-base)'s layer-wise performance with different pooling methods on the STS-B test set.
We observe that the performance can be dramatically varied according to the selected layer and pooling strategy.
Our self-guided training (SG / SG-OPT) assures much improved results compared to those of the baselines. } 
\label{fig:figure1}
\end{center}
\end{figure}

On the other hand, in cases where labeled datasets are unavailable, it is unclear what the best strategy is for deriving sentence embeddings from BERT.\footnote{In this paper, the term \textit{BERT} has two meanings: Narrowly, the BERT model itself, and more broadly, pre-trained Transformer encoders that share the same spirit with BERT.}
In practice, previous studies \cite{reimers-gurevych-2019-sentence,li-etal-2020-sentence,hu-et-al-2020-xtreme} reported that na\"ively (i.e., without any processing) leveraging the \verb|[CLS]| embedding as a sentence representation, as is the case of supervised fine-tuning, results in disappointing outcomes.
Currently, the most common rule of thumb for building BERT sentence embeddings without supervision is to apply mean pooling on the last layer(s) of BERT. 
Yet, this approach can be still sub-optimal.
In a preliminary experiment, we constructed sentence embeddings by employing various combinations of different BERT layers and pooling methods, and tested them on the Semantic Textual Similarity (STS) benchmark dataset \cite{cer-etal-2017-semeval}.\footnote{In the experiment, we employ the settings identical with ones used in Chapter \ref{sec:experiments}. Refer to Chapter \ref{sec:experiments} for more details.} 
We discovered that BERT(-base)'s performance, measured in Spearman correlation ($\times$ 100), can range from as low as 16.71 (\verb|[CLS]|, the 10$^{\text{th}}$ layer) to 63.19 (max pooling, the 2$^{\text{nd}}$ layer) depending on the selected layer and pooling method (see Figure \ref{fig:figure1}).
This result suggests that the current practice of building BERT sentence vectors is not solid enough, and that there is room to bring out more of BERT's expressiveness.

In this work, we propose a contrastive learning method that makes use of a newly proposed self-guidance mechanism to tackle the aforementioned problem.
The core idea is to recycle intermediate BERT hidden representations as positive samples to which the final sentence embedding should be close.
As our method does not require data augmentation, which is essential in most recent contrastive learning frameworks, it is much simpler and easier to use than existing methods \cite{fang2020cert,xie2020unsupervised}.
Moreover, we customize the NT-Xent loss \cite{chen-etal-2020-a}, a contrastive learning objective widely used in computer vision, for better sentence representation learning with BERT. 
We demonstrate that our approach outperforms competitive baselines designed for building BERT sentence vectors \cite{li-etal-2020-sentence,wang2020sbert} in various environments.
With comprehensive analyses, we also show that our method is more computationally efficient than the baselines at inference in addition to being more robust to domain shifts.

\section{Related Work}

\paragraph{Contrastive Representation Learning.}
Contrastive learning has been long considered as effective in constructing meaningful representations.
For instance, \citet{mikolov2013distributed} propose to learn word embeddings by framing words nearby a target word as positive samples while others as negative.
\citet{logeswaran2018an} generalize the approach of \citet{mikolov2013distributed} for sentence representation learning.
More recently, several studies \cite{fang2020cert,giorgi2020declutr,wu2020clear} suggest to utilize contrastive learning for training Transformer models, similar to our approach.
However, they generally require data augmentation techniques, e.g., back-translation \cite{sennrich2016improving}, or prior knowledge on training data such as order information, while our method does not.
Furthermore, we focus on revising BERT for computing better sentence embeddings rather than training a language model from scratch.

On the other hand, contrastive learning has been also receiving much attention from the computer vision community (\citet{chen-etal-2020-a,chen2020exploring,he2020momentum}, \textit{inter alia}).
We improve the framework of \citet{chen-etal-2020-a} by optimizing its learning objective for pre-trained Transformer-based sentence representation learning.
For extensive surveys on contrastive learning, refer to \citet{le2020contrastive} and \citet{jaiswal2020survey}.

\paragraph{Fine-tuning BERT with Supervision.}

It is not always trivial to fine-tune pre-trained Transformer models of gigantic size with success, especially when the number of target domain data is limited \cite{mosbach2020stability}.
To mitigate this training instability problem, several approaches \cite{aghajanyan2020better,jiang-etal-2020-smart,Zhu2020FreeLB} have been recently proposed.
In particular, \citet{gunel2021supervised} propose to exploit contrastive learning as an auxiliary training objective during fine-tuning BERT with supervision from target tasks.
In contrast, we deal with the problem of adjusting BERT when such supervision is not available.

\paragraph{Sentence Embeddings from BERT.}

Since BERT and its variants are originally designed to be fine-tuned on each downstream task to attain their optimal performance, it remains ambiguous how best to extract general sentence representations from them, which are broadly applicable across diverse sentence-related tasks.
Following \citet{conneau2017supervised}, \citet{reimers-gurevych-2019-sentence} (SBERT) propose to compute sentence embeddings by conducting mean pooling on the last layer of BERT and then fine-tuning the pooled vectors on the natural language inference (NLI) datasets \cite{bowman2015large,williams-etal-2018-broad}.
Meanwhile, some other studies concentrate on more effectively leveraging the knowledge embedded in BERT to construct sentence embeddings without supervision.
Specifically, \citet{wang2020sbert} propose a pooling method based on linear algebraic algorithms to draw sentence vectors from BERT's intermediate layers.
\citet{li-etal-2020-sentence} suggest to learn a mapping from the average of the embeddings obtained from the last two layers of BERT to a spherical Gaussian distribution using a flow model, and to leverage the redistributed embeddings in place of the original BERT representations.
We follow the setting of \citet{li-etal-2020-sentence} in that we only utilize plain text during training, however, unlike all the others that rely on a certain pooling method even after training, we directly refine BERT so that the typical \verb|[CLS]| vector can function as a sentence embedding.
Note also that there exists concurrent work \cite{carlsson2021semantic,gao2021simcse,wang2021tsdae} whose motivation is analogous to ours, attempting to improve BERT sentence embeddings in an unsupervised fashion.

\section{Method} \label{sec:method}

As BERT mostly requires some type of adaptation to be properly applied to a task of interest, it might not be desirable to derive sentence embeddings directly from BERT without fine-tuning.
While \citet{reimers-gurevych-2019-sentence} attempt to alleviate this problem with typical supervised fine-tuning, we restrict ourselves to revising BERT in an unsupervised manner, meaning that our method only demands a bunch of raw sentences for training.
  
Among possible unsupervised learning strategies, we concentrate on contrastive learning which can inherently motivate BERT to be aware of similarities between different sentence embeddings.
Considering that sentence vectors are widely used in computing the similarity of two sentences, the inductive bias introduced by contrastive learning can be helpful for BERT to work well on such tasks.
The problem is that sentence-level contrastive learning usually requires data augmentation \cite{fang2020cert} or prior knowledge on training data, e.g., order information \cite{logeswaran2018an}, to make plausible positive/negative samples.
We attempt to circumvent these constraints by utilizing the hidden representations of BERT, which are readily accessible, as samples in the embedding space.

\subsection{Contrastive Learning with Self-Guidance} \label{subsec:contrastive learning with self-guidance}

We aim at developing a contrastive learning method that is free from external procedure such as data augmentation.
A possible solution is to leverage (virtual) adversarial training \cite{miyato2018virtual} in the embedding space. 
However, there is no assurance that the semantics of a sentence embedding would remain unchanged when it is added with a random noise.
As an alternative, we propose to utilize the hidden representations from BERT's intermediate layers, which are conceptually guaranteed to represent corresponding sentences, as pivots that BERT sentence vectors should be close to or be away from.
We call our method as \textit{self-guided contrastive learning} since we exploit internal training signals made by BERT itself to fine-tune it.

\begin{figure}[t!]
\begin{center}
\includegraphics[width=\linewidth]{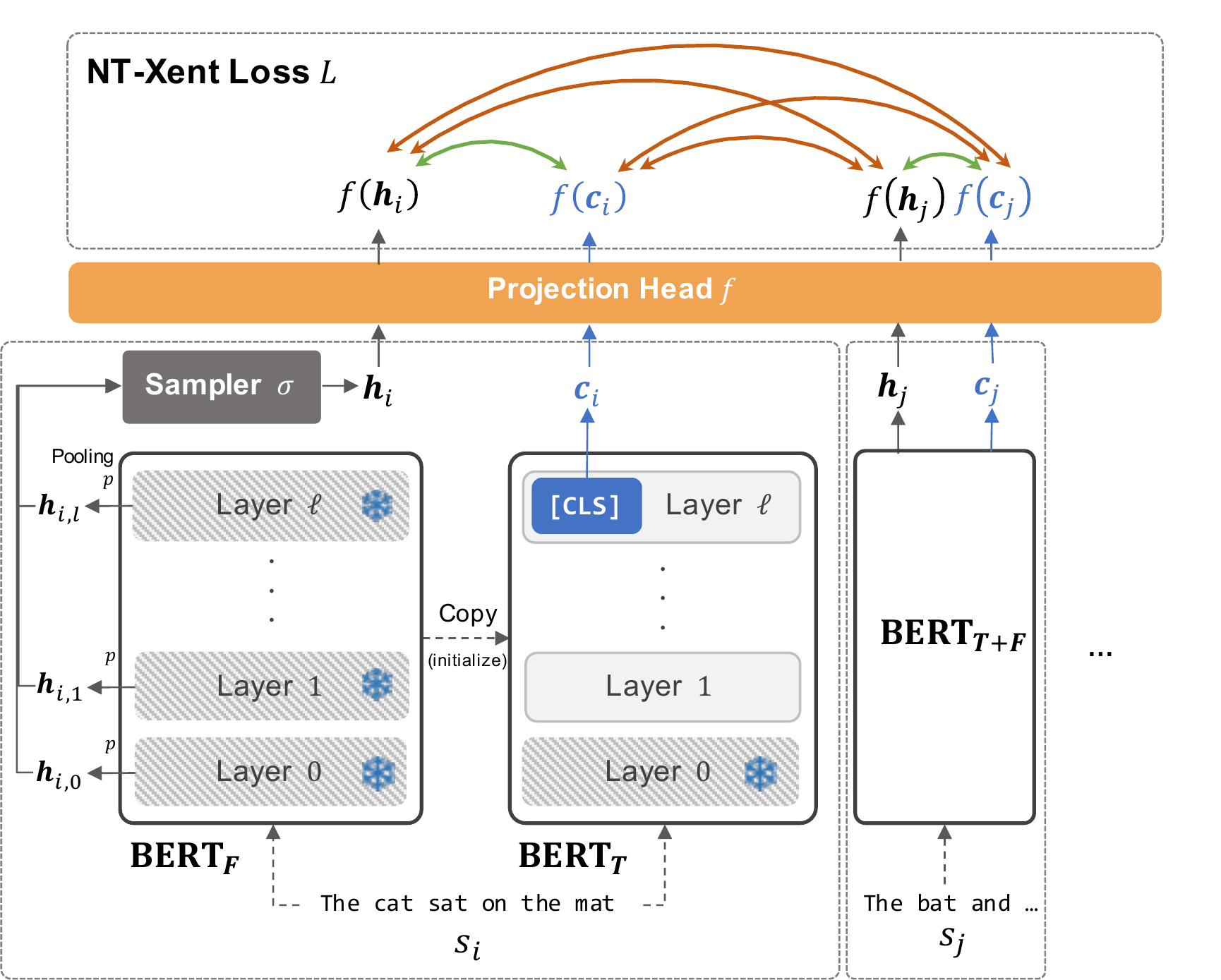}
\caption{Self-guided contrastive learning framework. We clone BERT into two copies at the beginning of training. $\text{BERT}_T$ (except Layer 0) is then fine-tuned to optimize the sentence vector $\mathbf{c}_i$ while $\text{BERT}_F$ is fixed.} 
\label{fig:figure2}
\end{center}
\end{figure}
 
We describe our training framework in Figure \ref{fig:figure2}. 
First, we clone BERT into two copies, $\text{BERT}_F$ (\textit{fixed}) and $\text{BERT}_T$ (\textit{tuned}) respectively.
$\text{BERT}_F$ is fixed during training to provide a training signal while $\text{BERT}_T$ is fine-tuned to construct better sentence embeddings.
The reason why we differentiate $\text{BERT}_F$ from $\text{BERT}_T$ is that we want to prevent the training signal computed by $\text{BERT}_F$ from being degenerated as the training procedure continues, which often happens when $\text{BERT}_F = \text{BERT}_T$.
This design decision also reflects our philosophy that our goal is to dynamically conflate the knowledge stored in BERT's different layers to produce sentence embeddings, rather than introducing new information via extra training.
Note that in our setting, the \verb|[CLS]| vector from the last layer of $\text{BERT}_T$, i.e., $\mathbf{c}_i$, is regarded as the final sentence embedding we aim to optimize/utilize during/after fine-tuning.

Second, given $b$ sentences in a mini-batch, say $s_1, s_2, \cdots, s_b$, 
we feed each sentence $s_i$ into $\text{BERT}_F$ and compute token-level hidden representations $H_{i,k} \in \mathbb{R}^{len(s_i) \times d} $:
\begin{equation*}
[H_{i,0}; H_{i,1};\cdots;H_{i,k};\cdots;H_{i,l}] = \text{BERT}_F(s_i),
\end{equation*}
where $ 0 \leq k \leq l $ (0: the non-contextualized layer), $l$ is the number of hidden layers in BERT, $ len(s_i)$ is the length of the tokenized sentence, and $d$ is the size of BERT's hidden representations.
Then, we apply a pooling function $p$ to $H_{i,k}$ for deriving diverse sentence-level views $ \mathbf{h}_{i,k} \in \mathbb{R}^{d}$ from all layers, i.e., $ \mathbf{h}_{i,k} = p(H_{i,k})$.
Finally, we choose the final view to be utilized by applying a sampling function $\sigma$:
\begin{equation*}
    \mathbf{h}_{i} = \sigma(\{\mathbf{h}_{i,k}|0 \leq k \leq l\}).
\end{equation*}
As we have no specific constraints in defining $p$ and $\sigma$, we employ max pooling as $p$ and a uniform sampler as $\sigma$ for simplicity, unless otherwise stated.
This simple choice for the sampler implies that each $\mathbf{h}_{i,k}$ has the same importance, which is persuasive considering it is known that different BERT layers are specialized at capturing disparate linguistic concepts \cite{jawahar-etal-2019-bert}.\footnote{We can also potentially make use of another sampler functions to inject our bias or prior knowledge on target tasks.}

Third, we compute our sentence embedding $\mathbf{c}_i$ for $s_i$ as follows:
\begin{equation*}
    \mathbf{c}_i = \text{BERT}_T(s_i)_{\text{[CLS]}},
\end{equation*}
where $\text{BERT}(\cdot)_{\text{[CLS]}}$ corresponds to the \verb|[CLS]| vector obtained from the last layer of BERT.
Next, we collect the set of the computed vectors into $ X  = \{\mathbf{x} | \mathbf{x} \in \{ \mathbf{c}_i \} \cup \{ \mathbf{h}_i \} \}$, and for all $\mathbf{x}_m \in X, $ we compute the NT-Xent loss \cite{chen-etal-2020-a}:
\begin{gather*}
    L_m^{base} = -\log{(\phi(\mathbf{x}_m, \mu(\mathbf{x}_m))/Z)}, \\
    \text{where } \phi(\mathbf{u}, \mathbf{v}) = \text{exp}(g(f(\mathbf{u}), f(\mathbf{v}))/\tau) \\
    \text{and } Z = \textstyle \sum_{n=1, n \neq m}^{2b} \phi(\mathbf{x}_m, \mathbf{x}_n).
\end{gather*}
Note that $\tau$ is a temperature hyperparameter, $f$ is a projection head consisting of MLP layers,\footnote{We employ a two-layered MLP whose hidden size is 4096. Each linear layer in the MLP is followed by a GELU function.} $ g(\mathbf{u},\mathbf{v}) = \mathbf{u} \cdot \mathbf{v} / \|\mathbf{u}\|\|\mathbf{v}\|$ is the cosine similarity function, and $\mu(\cdot)$ is the matching function defined as follows,
\begin{equation*}
    \mu(\mathbf{x}) =  
    \begin{cases}
        \mathbf{h}_i & \text{if } \mathbf{x} \text{ is equal to } \mathbf{c}_i. \\
        \mathbf{c}_i & \text{if } \mathbf{x} \text{ is equal to } \mathbf{h}_i.
    \end{cases}
\end{equation*}

Lastly, we sum all $L_m^{base}$ divided by $2b$, and add a regularizer $L^{reg} = \|\text{BERT}_{F}-\text{BERT}_{T}\|_2^2$ to prevent $\text{BERT}_T$ from being too distant from $\text{BERT}_F$.\footnote{To be specific, $L^{reg}$ is the square of the L2 norm of the difference between $\text{BERT}_F$ and $\text{BERT}_T$. As shown in Figure \ref{fig:figure2}, we also freeze the 0$^{th}$ layer of $\text{BERT}_T$ for stable learning.} 
As a result, the final loss $L^{base}$ is:
\begin{equation*}
    L^{base} = \frac{1}{2b} \sum_{m=1}^{2b} L_m^{base} + \lambda \cdot L^{reg},
\end{equation*}
where the coefficient $\lambda$ is a hyperparameter.

To summarize, our method refines BERT so that the sentence embedding $\mathbf{c}_i$ has a higher similarity with $\mathbf{h}_i$, which is another representation for the sentence $s_i$, in the subspace projected by $f$ while being relatively dissimilar with $\mathbf{c}_{j, j \neq i}$ and $\mathbf{h}_{j, j \neq i}$.
After training is completed, we remove all the components except $\text{BERT}_T$ and simply use $\mathbf{c}_i$ as the final sentence representation.

\begin{figure}[t!]
\begin{center}
\includegraphics[width=0.5\linewidth]{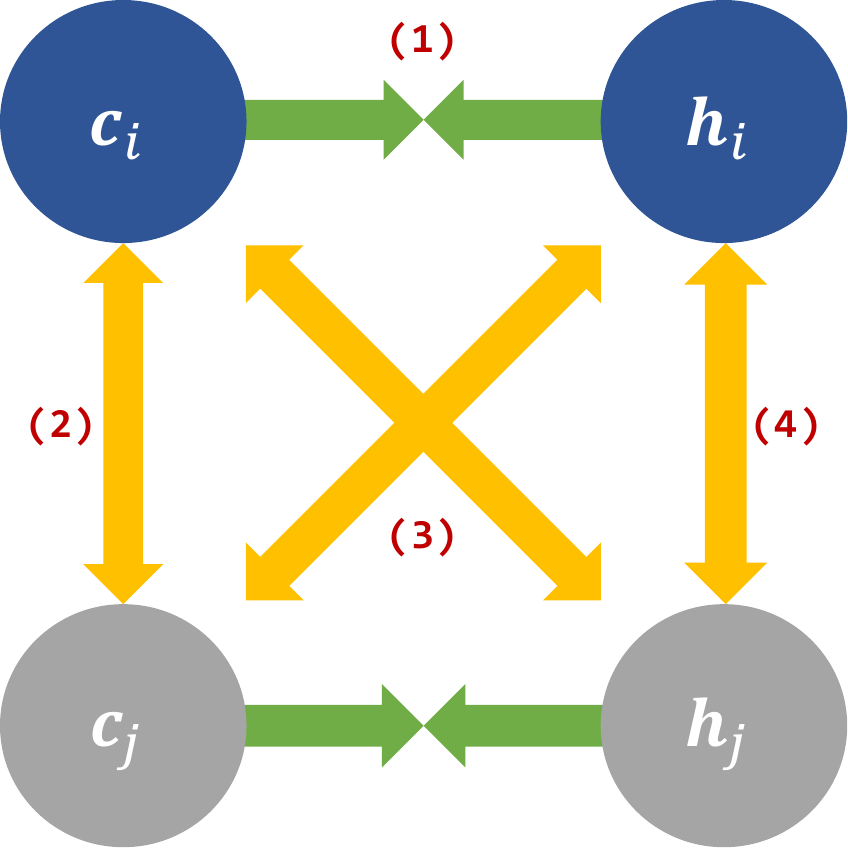}
\caption{Four factors of the original NT-Xent loss. Green and yellow arrows represent the force of attraction and repulsion, respectively. Best viewed in color.}
\label{fig:figure3}
\end{center}
\end{figure}

\subsection{Learning Objective Optimization} \label{subsec:learning objective optimization}

In Section \ref{subsec:contrastive learning with self-guidance}, we relied on a simple variation of the general NT-Xent loss, which is composed of four factors. Given sentence $s_i$ and $s_j$ without loss of generality, the factors are as follows (Figure \ref{fig:figure3}):
\begin{itemize}[noitemsep]
  \item[(1)] $\mathbf{c}_i \rightarrow\leftarrow \mathbf{h}_i$ (or $\mathbf{c}_j \rightarrow\leftarrow \mathbf{h}_j$): The main component that mirrors our core motivation that a BERT sentence vector ($\mathbf{c}_i$) should be consistent with intermediate views ($\mathbf{h}_i$) from BERT.
  \item[(2)] $\mathbf{c}_i \longleftrightarrow \mathbf{c}_j$: A factor that forces sentence embeddings ($\mathbf{c}_i$, $\mathbf{c}_j$) to be distant from each other.
  \item[(3)] $\mathbf{c}_i \longleftrightarrow \mathbf{h}_j$ (or $\mathbf{c}_j \longleftrightarrow \mathbf{h}_i$): An element that makes $\mathbf{c}_i$ being inconsistent with views for other sentences ($\mathbf{h}_j$).
 \item[(4)] $\mathbf{h}_i \longleftrightarrow \mathbf{h}_j$: A factor that causes a discrepancy between views of different sentences ($\mathbf{h}_i$, $\mathbf{h}_j$).
\end{itemize}
Even though all the four factors play a certain role, some components may be useless or even cause a negative influence on our goal.
For instance, \citet{chen2020exploring} have recently reported that in image representation learning, only (1) is vital while others are nonessential.
Likewise, we customize the training loss with three major modifications so that it can be more well-suited for our purpose.

First, as our aim is to improve $\mathbf{c}_i$ with the aid of $\mathbf{h}_i$, we re-define our loss focusing more on $\mathbf{c}_i$
rather than considering $\mathbf{c}_i$ and $\mathbf{h}_i$ as equivalent entities: 
\begin{gather*}
    L_i^{opt1} = -\log{(\phi(\mathbf{c}_i, \mathbf{h}_i) / \hat{Z})}, \\
    \text{where } \hat{Z} = \textstyle \sum_{j=1, j \neq i}^{b} \phi(\mathbf{c}_i, \mathbf{c}_j) + \sum_{j=1}^{b} \phi(\mathbf{c}_i, \mathbf{h}_j).
\end{gather*}
In other words, $\mathbf{h}_i$ only functions as points that $\mathbf{c}_i$ is encouraged to be close to or away from, and is not deemed as targets to be optimized.
This revision naturally results in removing (4).
Furthermore, we discover that (2) is also insignificant for improving performance, and thus derive $L_{i}^{opt2}$:
\begin{equation*}
    L_i^{opt2} = -\log({\phi(\mathbf{c}_i, \mathbf{h}_i)/ \textstyle \sum_{j=1}^{b} \phi(\mathbf{c}_i, \mathbf{h}_j)}).
\end{equation*}

Lastly, we diversify signals from (1) and (3) by allowing multiple views $\{\mathbf{h}_{i,k}\}$ to guide $\mathbf{c}_i$:
\begin{equation*}
    \resizebox{\linewidth}{!}{
    $L_{i,k}^{opt3} = -\log{\frac{\phi(\mathbf{c}_i, \mathbf{h}_{i,k})}{\phi(\mathbf{c}_i, \mathbf{h}_{i,k}) + \sum_{m=1, m \neq i}^{b}\sum_{n=0}^{l} \phi(\mathbf{c}_i, \mathbf{h}_{m,n})}}.$}
\end{equation*}
We expect with this refinement that the learning objective can provide more precise and fruitful training signals by considering additional (and freely available) samples being provided with.
The final form of our optimized loss is:
\begin{equation*}
    L^{opt} = \frac{1}{b(l+1)}\sum_{i=1}^{b} \sum_{k=0}^{l} L_{i,k}^{opt3} + \lambda \cdot L^{reg}.
\end{equation*}
In Section \ref{subsec:ablation study}, we show the decisions made in this section contribute to improvements in performance.

\section{Experiments} \label{sec:experiments}

\subsection{General Configurations} \label{subsec:general configurations}

In terms of pre-trained encoders, we leverage BERT \cite{devlin-etal-2019-bert} for English datasets and MBERT, which is a multilingual variant of BERT, for multilingual datasets.
We also employ RoBERTa \cite{liu2019roberta} and SBERT \cite{reimers-gurevych-2019-sentence} in some cases to evaluate the generalizability of tested methods.
We use the suffixes `-base' and `-large' to distinguish small and large models.
Every trainable model's performance is reported as the average of 8 separate runs to reduce randomness.
Hyperparameters are optimized on the STS-B validation set using BERT-base and utilized across different models.
See Table \ref{table:table8} in Appendix \ref{subsec:hyperparamters} for details.
Our implementation is based on the HuggingFace's \texttt{Transformers} \cite{wolf2019huggingface} and \texttt{SBERT} \cite{reimers-gurevych-2019-sentence} library, and publicly available at \url{https://github.com/galsang/SG-BERT}.

\begin{table*}[t!]
\scriptsize
    \centering
    \setlength{\tabcolsep}{0.6em}
    \begin{tabular}{l c c c c c c c c c}
    \toprule
    \textbf{Models} & \textbf{Pooling} & \textbf{STS-B} & \textbf{SICK-R} & \textbf{STS12} & \textbf{STS13} & \textbf{STS14} & \textbf{STS15} & \textbf{STS16} & \textbf{Avg.} \\ 
    \midrule
    \bf Non-BERT Baselines & & & & & & & & & \\ 
    GloVe$^{\dagger}$ & Mean & 58.02 & 53.76 & 55.14 & 70.66 & 59.73 & 68.25 & 63.66 & 61.32 \\
    USE$^{\dagger}$ & - & 74.92 & \underline{$76.69$} & 64.49 & 67.80 & 64.61 & 76.83 & 73.18 & 71.22 \\
    \midrule
    \bf BERT-base & & & & & & & & & \\ 
    + No tuning & CLS & 20.30 & 42.42 & 21.54 & 32.11 & 21.28 & 37.89 & 44.24 & 31.40 \\
    + No tuning & Mean & 47.29 & 58.22 & 30.87 & 59.89 & 47.73 & 60.29 & 63.73 & 52.57 \\
    + No tuning & WK & 16.07 & 41.54 & 16.01 & 21.80 & 15.96 & 33.59 & 34.07 & 25.58 \\
    + Flow & Mean-2 & 71.35$_{\pm0.27}$ & 64.95$_{\pm0.16}$ & 64.32$_{\pm0.17}$ & 69.72$_{\pm0.25}$ & 63.67$_{\pm0.06}$ & 77.77$_{\pm0.15}$ & 69.59$_{\pm0.28}$ & 68.77$_{\pm0.07}$ \\
    + Contrastive (BT) & CLS & 63.27$_{\pm1.48}$ & 66.91$_{\pm1.29}$ & 54.26$_{\pm1.84}$ & 64.03$_{\pm2.35}$ & 54.28$_{\pm1.87}$ & 68.19$_{\pm0.95}$ & 67.50$_{\pm0.96}$ & 62.63$_{\pm1.28}$ \\
    + \textbf{Contrastive (SG)} & CLS & 75.08$_{\pm0.73}$ & \textbf{68.19}$_{\pm0.36}$ & 63.60$_{\pm0.98}$ & 76.48$_{\pm0.69}$ & 67.57$_{\pm0.57}$ & 79.42$_{\pm0.49}$ & 74.85$_{\pm0.54}$ & 72.17$_{\pm0.44}$ \\
    + \textbf{Contrastive (SG-OPT)} & CLS & \textbf{77.23}$_{\pm0.43}$ & 68.16$_{\pm0.50}$ & \textbf{66.84}$_{\pm0.73}$ & \textbf{80.13}$_{\pm0.51}$ & \textbf{71.23}$_{\pm0.40}$ & \textbf{81.56}$_{\pm0.28}$ & \textbf{77.17}$_{\pm0.22}$ & \textbf{74.62}$_{\pm0.25}$ \\
    \midrule
    \bf BERT-large & & & & & & & & & \\
    + No tuning & CLS & 26.75 & 43.44 & 27.44 & 30.76 & 22.59 & 29.98 & 42.74 & 31.96 \\
    + No tuning & Mean & 47.00 & 53.85 & 27.67 & 55.79 & 44.49 & 51.67 & 61.88 & 48.91 \\
    + No tuning & WK & 35.75 & 38.39 & 12.65 & 26.41 & 23.74 & 29.34 & 34.42 & 28.67 \\
    + Flow & Mean-2 & 72.72$_{\pm0.36}$ & 63.77$_{\pm0.18}$ & 62.82$_{\pm0.17}$ & 71.24$_{\pm0.22}$ & 65.39$_{\pm0.15}$ & 78.98$_{\pm0.21}$ & 73.23$_{\pm0.24}$ & 70.07$_{\pm0.81}$ \\
    + Contrastive (BT) & CLS & 63.84$_{\pm1.05}$ & 66.53$_{\pm2.62}$ & 52.04$_{\pm1.75}$ & 62.59$_{\pm1.84}$ & 54.25$_{\pm1.45}$ & 71.07$_{\pm1.11}$ & 66.71$_{\pm1.08}$ & 62.43$_{\pm1.07}$ \\
    + \textbf{Contrastive (SG)} & CLS & 75.22$_{\pm0.57}$ & 69.63$_{\pm0.95}$ & 64.37$_{\pm0.72}$ & 77.59$_{\pm1.01}$ & 68.27$_{\pm0.40}$ & 80.08$_{\pm0.28}$ & 74.53$_{\pm0.43}$ & 72.81$_{\pm0.31}$ \\
    + \textbf{Contrastive (SG-OPT)} & CLS & \textbf{76.16}$_{\pm0.42}$ & \textbf{70.20}$_{\pm0.65}$ & \textbf{67.02}$_{\pm0.72}$ & \textbf{79.42}$_{\pm0.80}$ & \textbf{70.38}$_{\pm0.65}$ & \textbf{81.72}$_{\pm0.32}$ & \textbf{76.35}$_{\pm0.22}$ &	\textbf{74.46}$_{\pm0.35}$ \\								
    \midrule
    \bf SBERT-base & & & & & & & & & \\ 
    + No tuning & CLS & 73.66 & 69.71 & 70.15 & 71.17 & 68.89 & 75.53 & 70.16 & 71.32 \\
    + No tuning & Mean & 76.98 & 72.91 & 70.97 & 76.53 & 73.19 & 79.09 & 74.30 & 74.85 \\
    + No tuning & WK & 78.38 & 74.31 & 69.75 & 76.92 & 72.32 & 81.17 & 76.25 & 75.59 \\
    + Flow$^{\ddagger}$ & Mean-2 & 81.03 & 74.97 & 68.95 & 78.48 & \textbf{77.62} & 81.95 & 78.94 & 77.42 \\
    + Contrastive (BT) & CLS & 74.67$_{\pm0.30}$ & 70.31$_{\pm0.45}$ & 71.19$_{\pm0.37}$ & 72.41$_{\pm0.60}$ & 69.90$_{\pm0.43}$ & 77.16$_{\pm0.48}$ & 71.63$_{\pm0.55}$ & 72.47$_{\pm0.37}$ \\
    + \textbf{Contrastive (SG)} & CLS & 81.05$_{\pm0.34}$ & 75.78$_{\pm0.55}$ & 73.76$_{\pm0.76}$ & 80.08$_{\pm0.45}$ & 75.58$_{\pm0.57}$ & 83.52$_{\pm0.43}$ & 79.10$_{\pm0.51}$ & 78.41$_{\pm0.33}$ \\
    + \textbf{Contrastive (SG-OPT)} & CLS & \textbf{81.46}$_{\pm0.27}$ & \textbf{76.64} $_{\pm0.42}$ & \underline{\textbf{75.16}}$_{\pm0.56}$ & \textbf{81.27}$_{\pm0.37}$ & 76.31$_{\pm0.38}$ & \underline{\textbf{84.71}}$_{\pm0.26}$ & \textbf{80.33}$_{\pm0.19}$ & \textbf{79.41}$_{\pm0.17}$ \\
    \midrule
    \bf SBERT-large & & & & & & & & & \\
    + No tuning & CLS & 76.01 & 70.99 & 69.05 & 71.34 & 69.50 & 76.66 & 70.08 & 71.95 \\
    + No tuning & Mean & 79.19 & 73.75 & 72.27 & 78.46 & 74.90 & 80.99 & 76.25 & 76.54 \\
    + No tuning & WK & 61.87 & 67.06 & 49.95 & 53.02 & 46.55 & 62.47 & 60.32 & 57.32 \\
    + Flow$^{\ddagger}$ & Mean-2 & 81.18 & 74.52 & 70.19 & 80.27 & \underline{\textbf{78.85}} & 82.97 & \underline{\textbf{80.57}} & 78.36 \\
    + Contrastive (BT) & CLS & 76.71$_{\pm1.22}$ & 71.56$_{\pm1.34}$ & 69.95$_{\pm3.57}$ & 72.66$_{\pm1.16}$ & 70.38$_{\pm2.10}$ & 77.80$_{\pm3.24}$ & 71.41$_{\pm1.73}$ & 72.92$_{\pm1.53}$ \\
    + \textbf{Contrastive (SG)} & CLS & \underline{\textbf{82.35}}$_{\pm0.15}$ & \textbf{76.44}$_{\pm0.41}$ & \textbf{74.84}$_{\pm0.57}$ & \textbf{82.89}$_{\pm0.41}$ & 77.27$_{\pm0.35}$ & 84.44$_{\pm0.23}$ & 79.54$_{\pm0.49}$ & 79.68$_{\pm0.37}$ \\
    + \textbf{Contrastive (SG-OPT)} & CLS & 82.05$_{\pm0.39}$ & \textbf{76.44}$_{\pm0.29}$ & 74.58$_{\pm0.59}$ & \underline{\textbf{83.79}}$_{\pm0.14}$ & 76.98$_{\pm0.19}$ & \textbf{84.57}$_{\pm0.27}$ & 79.87$_{\pm0.42}$ & \underline{\textbf{79.76}}$_{\pm0.33}$ \\
    \bottomrule
    \end{tabular}
    \caption{Experimental results on STS tasks. 
    Results for trained models are averaged over 8 runs ($\pm$: the standard deviation).
    The best figure in each (model-wise) part is in \textbf{bold} and the best in each column is \underline{underlined}.
    Our method with self-guidance (SG, SG-OPT) generally outperforms competitive baselines. 
    We borrow scores from previous work if we could not reproduce them. 
    $\dagger$: from \citet{reimers-gurevych-2019-sentence}. $\ddagger$: from \citet{li-etal-2020-sentence}.} 
    \label{table:table1}
\end{table*}

\subsection{Semantic Textual Similarity Tasks} \label{subsec:semantic textual similairty tasks}

We first evaluate our method and baselines on Semantic Textual Similarity (STS) tasks.
Given two sentences, we derive their similarity score by computing the cosine similarity of their embeddings.

\paragraph{Datasets and Metrics.} 
Following the literature, we evaluate models on 7 datasets in total, that is, STS-B \cite{cer-etal-2017-semeval}, SICK-R \cite{marelli-etal-2014-sick}, and STS12-16 \cite{agirre-etal-2012-semeval,agirre-etal-2013-sem,agirre2014semeval,agirre-etal-2015-semeval,agirre-etal-2016-semeval}.
These datasets contain pairs of two sentences,  whose similarity scores are labeled from 0 to 5.
The relevance between gold annotations and the scores predicted by sentence vectors is measured in Spearman correlation ($\times$ 100).

\paragraph{Baselines and Model Specification.} 
We first prepare two non-BERT approaches as baselines, i.e., Glove \cite{pennington2014glove} mean embeddings and Universal Sentence Encoder (USE; \citet{cer2018universal}).
In addition, various methods for BERT sentence embeddings that do not require supervision are also introduced as baselines:

\begin{itemize}[noitemsep,leftmargin=*]
    \item \textbf{CLS} token embedding: It regards the \verb|[CLS]| vector from the last layer of BERT as a sentence representation.
    \item \textbf{Mean} pooling: This method conducts mean pooling on the last layer of BERT and use the output as a sentence embedding.
    \item \textbf{WK} pooling: This follows the method of \citet{wang2020sbert}, which exploits QR decomposition and extra techniques to derive meaningful sentence vectors from BERT.
    \item \textbf{Flow}: This is \textit{BERT-flow} proposed by \citet{li-etal-2020-sentence}, which is a flow-based model that maps the vectors made by taking mean pooling on the last two layers of BERT to a Gaussian space.\footnote{We restrictively utilize this model, as we find it difficult to exactly reproduce the model's result with its official code.}
    \item \textbf{Contrastive (BT)}: Following \citet{fang2020cert}, we revise BERT with contrastive learning. However, this method relies on back-translation to obtain positive samples, unlike ours. Details about this baseline are specified in Appendix \ref{subsec:specification on Contrastive (BT)}. 
\end{itemize}
We make use of plain sentences from STS-B to fine-tune BERT using our approach, identical with Flow.\footnote{For training, \citet{li-etal-2020-sentence} utilize the concatenation of the STS-B training, validation, and test set (\textit{without} gold annotations). 
We also follow the same setting for a fair comparison.}
We name the BERT instances trained with our self-guided method as \textbf{Contrastive (SG)} and \textbf{Contrastive (SG-OPT)}, which utilize $L^{base}$ and $L^{opt}$ in Section \ref{sec:method} respectively.

\paragraph{Results.} 

We report the performance of different approaches on STS tasks in Table \ref{table:table1} and Table \ref{table:table11} (Appendix \ref{subsec:roberta's performance on sts tasks}).
From the results, we confirm the fact that our methods (SG and SG-OPT) mostly outperform other baselines in a variety of experimental settings.
As reported in earlier studies, the na\"ive \verb|[CLS]| embedding and mean pooling are turned out to be inferior to sophisticated methods.
To our surprise, WK pooling's performance is even lower than that of mean pooling in most cases, and the only exception is when WK pooling is applied to SBERT-base.
Flow shows its strength outperforming the simple strategies. 
Nevertheless, its performance is shown to be worse than that of our methods (although some exceptions exist in the case of SBERT-large).
Note that contrastive learning becomes much more competitive when it is combined with our self-guidance algorithm rather than back-translation.
It is also worth mentioning that the optimized version of our method (SG-OPT) generally shows better performance than the basic one (SG), proving the efficacy of learning objective optimization (Section \ref{subsec:learning objective optimization}).
To conclude, we demonstrate that our self-guided contrastive learning is effective in improving the quality of BERT sentence embeddings when tested on STS tasks.

\begin{table}[t!]
\scriptsize
    \centering
    \begin{tabular}{l c}
    \toprule
    \textbf{Models} & \textbf{Spanish} \\
    \midrule
    \textbf{Baseline} \cite{agirre2014semeval} & \\ 
    UMCC-DLSI-run2 (Rank \#1) & 80.69 \\
    \midrule
    \textbf{MBERT} & \\
    + CLS & 12.60 \\
    + Mean pooling & 81.14 \\
    + WK pooling & 79.78 \\
    + Contrastive (BT) & 78.04 \\
    + \textbf{Contrastive (SG)} & 82.09 \\
    + \textbf{Contrastive (SG-OPT)} & 82.74 \\
    \bottomrule
    \end{tabular}
    \caption{SemEval-2014 Task 10 Spanish task.} 
    \label{table:table2}
\end{table}

\begin{table}[t!]
\scriptsize 
    \centering
    \setlength{\tabcolsep}{0.5em}
    \begin{tabular}{l c c c}
    \toprule
    \multirow{2}{*}{\textbf{Models}} & \textbf{Arabic} & \textbf{Spanish} & \textbf{English} \\ 
    & \textbf{(Track 1)} & \textbf{(Track 3)} & \textbf{(Track 5)} \\
    \midrule
    \textbf{Baselines} & & & \\
    Cosine baseline	\cite{cer-etal-2017-semeval} & 60.45 & 71.17 & 72.78 \\
    ENCU (Rank \#1, \citet{tian-etal-2017-ecnu}) & 74.40 & 85.59 & 85.18 \\
    \midrule
    \textbf{MBERT} & & & \\
    + CLS & 30.57 & 29.38 & 24.97 \\
    + Mean pooling & 51.09 & 54.56 & 54.86 \\
    + WK pooling & 50.38 & 55.87 & 54.87  \\
    + Contrastive (BT) & 54.24 & 68.16 & 73.89 \\
    \textbf{+ Contrastive (SG)} & 57.09 & 78.93 & 78.24 \\
    + \textbf{Contrastive (SG-OPT)} & 58.52 & 80.19 & 78.03 \\
    \bottomrule
    \end{tabular}
    \caption{Results on SemEval-2017 Task 1: Track 1 (Arabic), Track 3 (Spanish), and Track 5 (English).}
    \label{table:table3}
\end{table}

\subsection{Multilingual STS Tasks}

We expand our experiments to multilingual settings by utilizing MBERT and cross-lingual zero-shot transfer.
Specifically, we refine MBERT using only English data and test it on datasets written in other languages.
As in Section \ref{subsec:semantic textual similairty tasks}, we use the English STS-B for training.
We consider two datasets for evaluation: (1) SemEval-2014 Task 10 (Spanish; \citet{agirre2014semeval}) and (2) SemEval-2017 Task 1   (Arabic, Spanish, and English; \citet{cer-etal-2017-semeval}).
Performance is measured in Pearson correlation ($\times$ 100) for a fair comparison with previous work.

From Table \ref{table:table2}, we see that MBERT with mean pooling already outperforms the best system (at the time of the competition was held) on SemEval-2014 and that our method further boosts the model's performance.
In contrast, in the case of SemEval-2017 (Table \ref{table:table3}), MBERT with mean pooling even fails to beat the strong Cosine baseline.\footnote{The Cosine baseline computes its score as the cosine similarity of binary sentence vectors with each dimension representing whether an individual word appears in a sentence.} 
However, MBERT becomes capable of outperforming (in English/Spanish) or being comparable with (Arabic) the baseline by adopting our algorithm.
We observe that while cross-lingual transfer using MBERT looks promising for the languages analogous to English (e.g., Spanish), its effectiveness may shrink on distant languages (e.g., Arabic).
Compared against the best system which is trained on task-specific data, MBERT shows reasonable performance considering that it is never exposed to any labeled STS datasets.
In summary, we demonstrate that MBERT fine-tuned with our method has a potential to be used as a simple but effective tool for multilingual (especially European) STS tasks.

\begin{table}[t!]
\scriptsize
    \centering
    \setlength{\tabcolsep}{0.25em}
    \begin{tabular}{l c c c c c c c c }
    \toprule
    \textbf{Models} & \textbf{MR} & \textbf{CR} & \textbf{SUBJ} & \textbf{MPQA} & \textbf{SST2} & \textbf{TREC} & \textbf{MRPC} & \textbf{Avg.} \\ 
    \midrule
    \bf BERT-base & & & & & & & & \\ 
    + Mean & 81.46 & 86.71 & 95.37 & 87.90 & 85.83 & 90.30 & 73.36 & 85.85 \\
    + WK & 80.64 & 85.53 & 95.27 & 88.63 & 85.03 & \textbf{94.03} & 71.71 & 85.83 \\
    + \textbf{SG-OPT} & \textbf{82.47} & \textbf{87.42} & \textbf{95.40} & \textbf{88.92} & \textbf{86.20} & 91.60 & \textbf{74.21} &
    \textbf{86.60} \\
    \midrule
    \bf BERT-large & & & & & & & & \\
    + Mean & 84.38 & 89.01 & 95.60 & 86.69 & 89.20 & 90.90 & 72.79 & 86.94 \\
    + WK & 82.68 & 87.92 & 95.32 & \textbf{87.25} & 87.81 & 91.18 & 70.13 & 86.04 \\
    + \textbf{SG-OPT} & \textbf{86.03} & \textbf{90.18} & \textbf{95.82} & 87.08 & \textbf{90.73} & \textbf{94.65} & \textbf{73.31} & \textbf{88.26} \\
    \midrule 
    \bf SBERT-base & & & & & & & & \\
    + Mean & 82.80 & 89.03 & 94.07 & 89.79 & 88.08 & 86.93 & 75.11 & 86.54 \\
    + WK & 82.96 & 89.33 & \textbf{95.13} & \textbf{90.56} & 88.10 & \textbf{91.98} & \textbf{76.66} & \textbf{87.82} \\
    + \textbf{SG-OPT} & \textbf{83.34} & \textbf{89.45} & 94.68 & 89.78 & \textbf{88.57} & 87.30 & 75.26 & 86.91 \\
    \bottomrule
    \end{tabular}
    \caption{Experimental results on SentEval.}
    \label{table:table4}
\end{table}

\subsection{SentEval and Supervised Fine-tuning} \label{subsec:senteval and supervised fine-tuning}

We also evaluate BERT sentence vectors using the SentEval \cite{conneau-kiela-2018-senteval} toolkit.
Given sentence embeddings, SentEval trains linear classifiers on top of them and estimates the quality of the vectors via their performance (accuracy) on downstream tasks.
Among available tasks, we employ 7: MR, CR, SUBJ, MPQA, SST2, TREC, MRPC.\footnote{Refer to \citet{conneau-kiela-2018-senteval} for each task's spec.}

In Table \ref{table:table4}, we compare our method (SG-OPT) with two baselines.\footnote{
We focus on reporting our own results as we discovered that the toolkit's outcomes can be fluctuating depending on its configuration (we list our settings in Appendix \ref{subsec:senteval configurations}). 
We also restrict ourselves to evaluating SG-OPT for simplicity, as SG-OPT consistently showed better performance than other contrastive methods in previous experiments.}
We find that our method is helpful over usual mean pooling in improving the performance of BERT-like models on SentEval.
SG-OPT also outperforms WK pooling on BERT-base/large while being comparable on SBERT-base.
From the results, we conjecture that self-guided contrastive learning and SBERT training suggest a similar inductive bias in a sense, as the benefit we earn by revising SBERT with our method is relatively lower than the gain we obtain when fine-tuning BERT.
Meanwhile, it seems that WK pooling provides an orthogonal contribution that is effective in the focused case, i.e., SBERT-base.

In addition, we examine how our algorithm impacts on supervised fine-tuning of BERT, although it is not the main concern of this work.
Briefly reporting, we identify that the original BERT(-base) and one tuned with SG-OPT show comparable performance on the GLUE \cite{wang2019glue} validation set, implying that our method does not influence much on BERT's supervised fine-tuning.
We refer readers to Appendix \ref{subsec:glue experiments} for more details.

\section{Analysis}

We here further investigate the working mechanism of our method with supplementary experiments.
All the experiments conducted in this section follow the configurations stipulated in Section \ref{subsec:general configurations} and \ref{subsec:semantic textual similairty tasks}.

\begin{table}[t!]
\scriptsize
    \centering
    \setlength{\tabcolsep}{0.2em}
    \begin{tabular}{l c}
    \toprule
    \textbf{Models} & \textbf{STS Tasks (Avg.)} \\ 
    \midrule
    \textbf{BERT-base} \\
    + \textbf{SG-OPT} ($L^{opt3}$) & 74.62 \\
    + $L^{opt2}$ & 73.14 (-1.48) \\
    + $L^{opt1}$ & 72.61 (-2.01) \\
    + \textbf{SG} ($L^{base}$) & 72.17 (-2.45) \\
    \midrule 
    \textbf{BERT-base} + \textbf{SG-OPT} ($\tau=0.01$, $\lambda=0.1$) & 74.62 \\ 
    + $\tau=0.1$ & 70.39 (-4.23)\\
    + $\tau=0.001$ & 74.16 (-0.46) \\
    + $\lambda=0.0$ & 73.76 (-0.86) \\ 
    + $\lambda=1.0$ & 73.18 (-1.44) \\ 
    - Projection head ($f$) & 72.78 (-1.84) \\
    \bottomrule
    \end{tabular}
    \caption{Ablation study.}
    \label{table:table5}
\end{table}

\subsection{Ablation Study} \label{subsec:ablation study}

We conduct an ablation study to justify the decisions made in optimizing our algorithm.
To this end, we evaluate each possible variant on the test sets of STS tasks.
From Table \ref{table:table5}, we confirm that all our modifications to the NT-Xent loss contribute to improvements in performance.
Moreover, we show that correct choices for hyperparameters are important for achieving the optimal performance, and that the projection head ($f$) plays a significant role as in \citet{chen-etal-2020-a}.

\begin{figure}[t!]
\begin{center}
\includegraphics[width=0.9\linewidth]{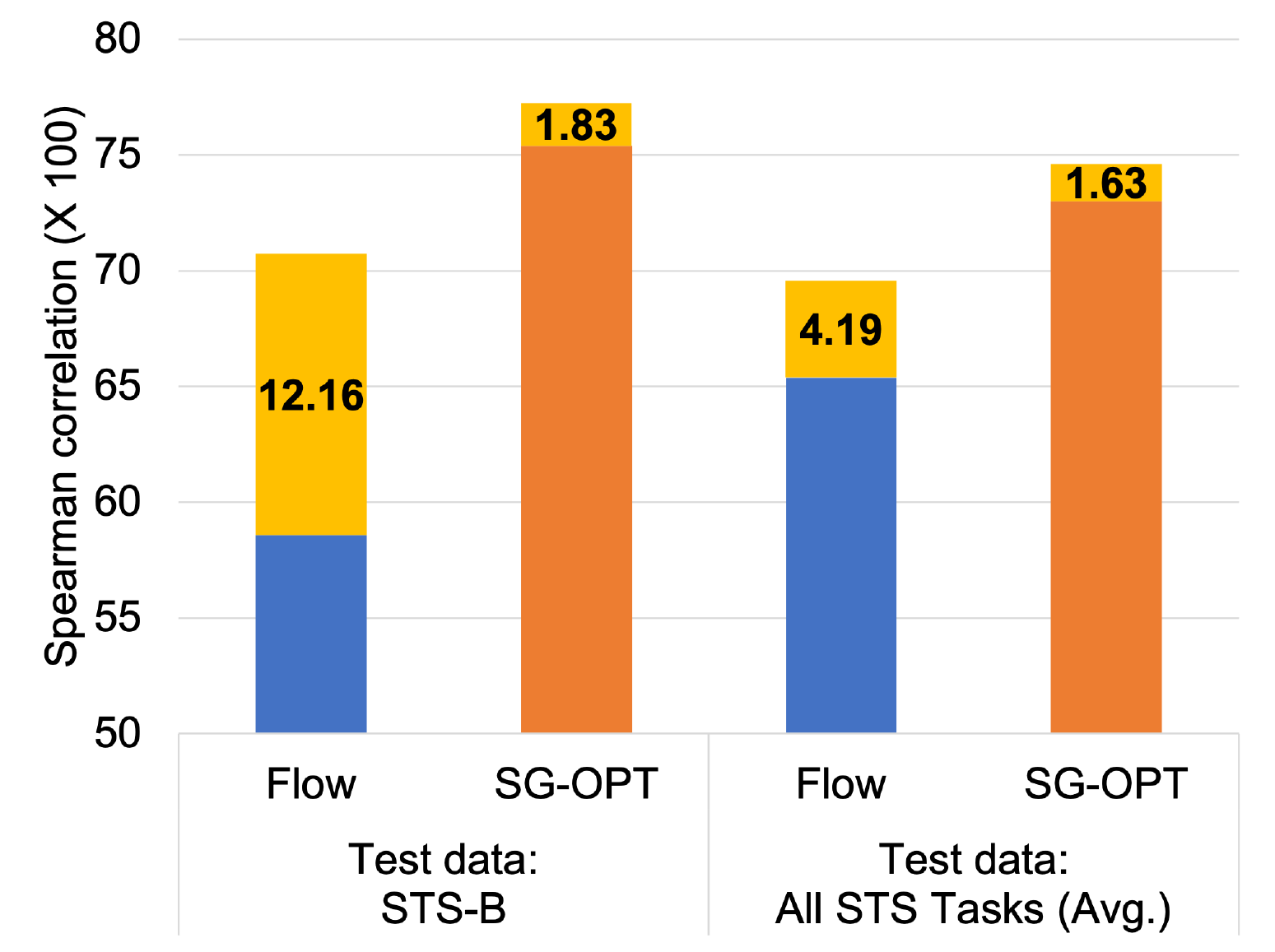}
\caption{Domain robustness study. 
The yellow bars indicate the performance gaps each method has according to which data it is trained with: in-domain (STS-B) or out-of-domain (NLI).
Our method (SG-OPT) clearly shows its relative robustness compared to Flow.} 
\label{fig:figure4}
\end{center}
\end{figure}

\subsection{Robustness to Domain Shifts}

Although our method in principle can accept any sentences in training, its performance might be varied with the training data it employs (especially depending on whether the training and test data share the same domain).
To explore this issue, we apply SG-OPT on BERT-base by leveraging the mix of NLI datasets \cite{bowman2015large,williams-etal-2018-broad} instead of STS-B, and observe the difference.
From Figure \ref{fig:figure4}, we confirm the fact that no matter which test set is utilized (STS-B or all the seven STS tasks), our method clearly outperforms Flow in every case, showing its relative robustness to domain shifts.
SG-OPT only loses 1.83 (on the STS-B test set) and 1.63 (on average when applied to all the STS tasks) points respectively when trained with NLI rather than STS-B, while Flow suffers from the considerable losses of 12.16 and 4.19 for each case.
Note, however, that follow-up experiments in more diverse conditions might be desired as future work, as the NLI dataset inherently shares some similarities with STS tasks.

\begin{table}[t!]
\scriptsize
    \centering
    \setlength{\tabcolsep}{0.2em}
    \begin{tabular}{l c c}
    \toprule
    \multirow{2}{*}[-0.4em]{\textbf{Layer}} & \multicolumn{2}{c}{\textbf{Elapsed Time}} \\
    \cmidrule(lr){2-3}
    & \textbf{Training (sec.)} & \textbf{Inference (sec.)} \\
    \midrule
    \textbf{BERT-base} & \\ 
    + Mean pooling & - & 13.94 \\ 
    + WK pooling & - & 197.03 ($\approx$ 3.3 min.) \\
    + Flow & 155.37 ($\approx$ 2.6 min.) & 28.49 \\ 
    + \textbf{Contrastive (SG-OPT)} & 455.02 ($\approx$ 7.5 min.) & 10.51 \\
    \bottomrule
    \end{tabular}
    \caption{Computational efficiency tested on STS-B.}
    \label{table:table6}
\end{table}

\subsection{Computational Efﬁciency}

In this part, we compare the computational efficiency of our method to that of other baselines. 
For each algorithm, we measure the time elapsed during training (if required) and inference when tested on STS-B.
All methods are run on the same machine (an Intel Xeon CPU E5-2620 v4 @ 2.10GHz and a Titan Xp GPU) using batch size 16.
The experimental results specified in Table \ref{table:table6} show that although our method demands a moderate amount of time ($<$ 8 min.) for training, it is the most efficient at inference, since our method is free from any post-processing such as pooling once training is completed.

\begin{figure}[t!]
\begin{center}
\includegraphics[width=0.9\linewidth]{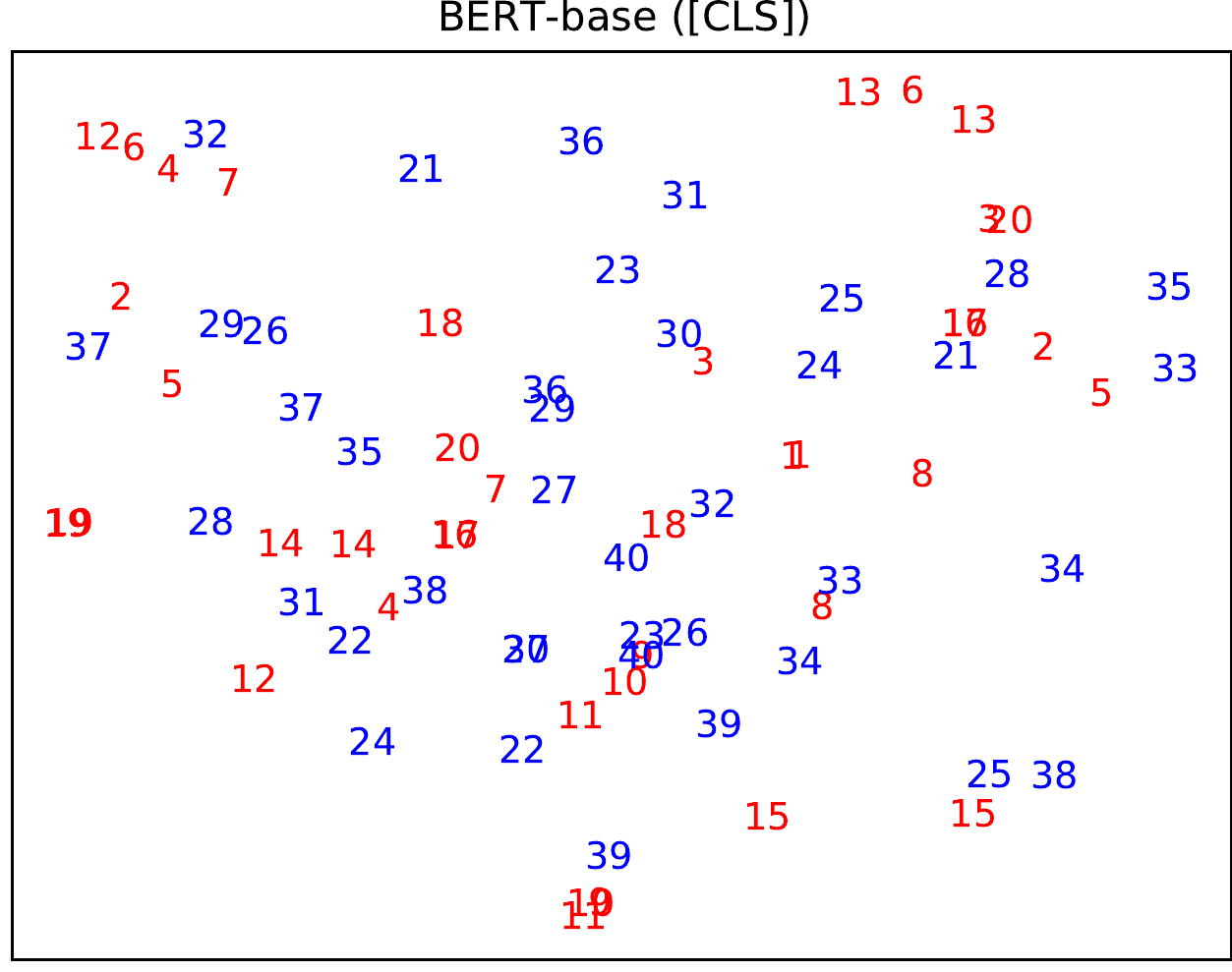}
\\ \vspace{0.2cm}
\includegraphics[width=0.9\linewidth]{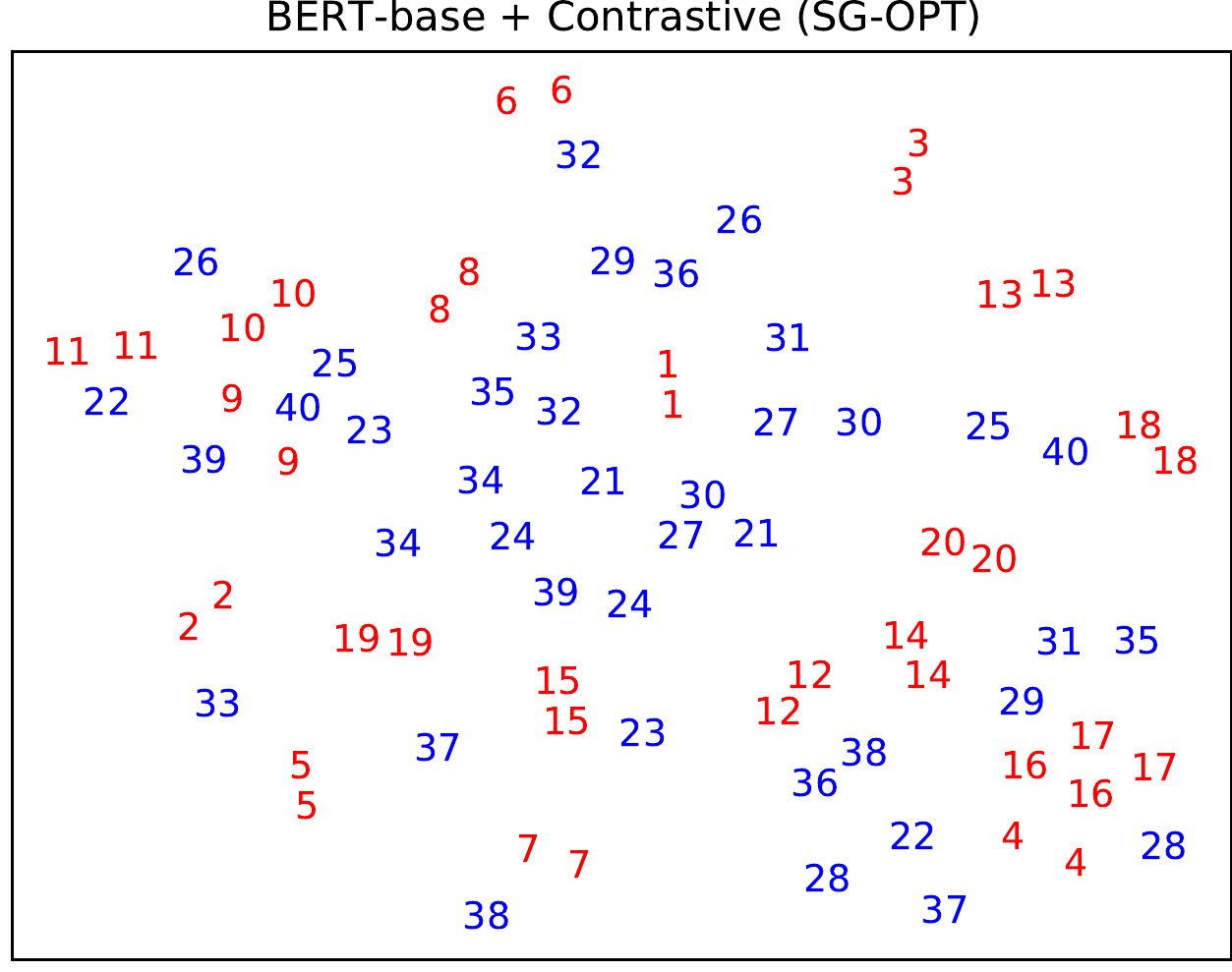}
\caption{Sentence representation visualization. (Top) Embeddings from the original BERT. (Bottom) Embeddings from the BERT instance fine-tuned with SG-OPT. Red numbers correspond to positive sentence pairs and blue to negative pairs.} 
\label{fig:figure5}
\end{center}
\end{figure}

\begin{table*}[t!]
\scriptsize
    \centering
    \setlength{\tabcolsep}{0.4em}
    \begin{tabular}{l c c c c c c c c c}
    \toprule
    \textbf{Models} & \textbf{Pooling} & \textbf{STS-B} & \textbf{SICK-R} & \textbf{STS12} & \textbf{STS13} & \textbf{STS14} & \textbf{STS15} & \textbf{STS16} & \textbf{Avg.} \\ 
    \midrule
    \bf BERT-base & & & & & & & & & \\
    + Contrastive (BT) & CLS & 63.27$_{\pm1.48}$ & 66.91$_{\pm1.29}$ & 54.26$_{\pm1.84}$ & 64.03$_{\pm2.35}$ & 54.28$_{\pm1.87}$ & 68.19$_{\pm0.95}$ & 67.50$_{\pm0.96}$ & 62.63$_{\pm1.28}$ \\
    + Contrastive (SG-OPT) & CLS & 77.23$_{\pm0.43}$ & 68.16$_{\pm0.50}$ & 66.84$_{\pm0.73}$ & \textbf{80.13}$_{\pm0.51}$ & 71.23$_{\pm0.40}$ & 81.56$_{\pm0.28}$ & 77.17$_{\pm0.22}$ & 74.62$_{\pm0.25}$ \\
    + \textbf{Contrastive (BT + SG-OPT)} & CLS & \textbf{77.99}$_{\pm0.23}$ & \textbf{68.75}$_{\pm0.79}$ & \textbf{68.49}$_{\pm0.38}$ & 80.00$_{\pm0.78}$ & \textbf{71.34}$_{\pm0.40}$ & \textbf{81.71}$_{\pm0.29}$ & \textbf{77.43}$_{\pm0.46}$ & \textbf{75.10}$_{\pm0.15}$ \\
    \bottomrule
    \end{tabular}
    \caption{Ensemble of the techniques for contrastive learning: back-translation (BT) and self-guidance (SG-OPT).}
    \label{table:table7}
\end{table*}

\subsection{Representation Visualization}

We visualize a few variants of BERT sentence representations to grasp an intuition on why our method is effective in improving performance.
Specifically, we sample 20 positive pairs (red, whose similarity scores are 5) and 20 negative pairs (blue, whose scores are 0) from the STS-B validation set. Then we compute their vectors and draw them on the 2D space with the aid of t-SNE.
In Figure \ref{fig:figure5}, we confirm that our SG-OPT encourages BERT sentence embeddings to be more well-aligned with their positive pairs while still being relatively far from their negative pairs.
We also visualize embeddings from SBERT (Figure \ref{fig:figure6} in Appendix \ref{subsec:representation visualization(SBERT)}), and identify that our approach and the supervised fine-tuning used in SBERT provide a similar effect, making the resulting embeddings more suitable for calculating correct similarities between them.

\section{Discussion}

In this section, we discuss a few weaknesses of our method in its current form and look into some possible avenues for future work.

First, while defining the proposed method in Section \ref{sec:method}, we have made decisions on some parts without much consideration about their optimality, prioritizing simplicity instead.
For instance, although we proposed utilizing all the intermediate layers of BERT and max pooling in a normal setting (indeed, it worked pretty well for most cases), a specific subset of the layers or another pooling method might bring better performance in a particular environment, as we observed in Section \ref{subsec:senteval and supervised fine-tuning} that we could achieve higher numbers by employing mean pooling and excluding lower layers in the case of SentEval (refer to Appendix \ref{subsec:senteval configurations} for details).
Therefore, in future work, it is encouraged to develop a systematic way of making more optimized design choices in specifying our method by considering the characteristics of target tasks.

Second, we expect that the effectiveness of contrastive learning in revising BERT can be improved further by properly combining different techniques developed for it.
As an initial attempt towards this direction, we conduct an extra experiment where we test the ensemble of back-translation and our self-guidance algorithm by inserting the original sentence into $\text{BERT}_T$ and its back-translation into $\text{BERT}_F$ when running our framework.
In Table \ref{table:table7}, we show that the fusion of the two techniques generally results in better performance, shedding some light on our future research direction.

\section{Conclusion}

In this paper, we have proposed a contrastive learning method with self-guidance for improving BERT sentence embeddings.
Through extensive experiments, we have demonstrated that our method can enjoy the benefit of contrastive learning without relying on external procedures such as data augmentation or back-translation, succeeding in generating higher-quality sentence representations compared to competitive baselines.
Furthermore, our method is efficient at inference because it does not require any post-processing once its training is completed, and is relatively robust to domain shifts.

\section*{Acknowledgments}

We would like to thank anonymous reviewers for their fruitful feedback.
We are also grateful to Jung-Woo Ha, Sang-Woo Lee, Gyuwan Kim, and other members in NAVER AI Lab in addition to Reinald Kim Amplayo for their insightful comments.

\bibliographystyle{acl_natbib}
\bibliography{acl2021}

\clearpage

\appendix

\section{Appendices} \label{sec:appendix}

\subsection{Hyperparameters} \label{subsec:hyperparamters}

\begin{table}[h]
\scriptsize
    \centering
    \begin{tabular}{l c}
    \toprule
    \textbf{Hyperparameters} & \textbf{Values} \\
    \midrule
    Random seed & 1, 2, 3, 4, 1234, 2345, 3456, 7890 \\
    Evaluation step & 50 \\ 
    Epoch & 1 \\ 
    Batch size ($b$) & 16 \\ 
    Optimizer & AdamW ($\beta_1, \beta_2$=(0.9, 0.9))\\
    Learning rate & 0.00005 \\ 
    Early stopping endurance & 10 \\
    $\tau$ & 0.01 \\
    $\lambda$ & 0.1 \\
    \bottomrule
    \end{tabular}
    \caption{Hyperparameters for experiments.}
    \label{table:table8}
\end{table}

\subsection{Specification on Contrastive (BT)} \label{subsec:specification on Contrastive (BT)}

This baseline is identical with our \textbf{Contrastive (SG)} model, except that it utilizes back-translation to generate positive samples.
To be specific, English sentences in the training set are traslated into German sentences using the WMT'19 English-German translator  provided by \citet{ng2019facebook}, and then the translated German sentences are back-translated into English with the aid of the WMT'19 German-English model also offered by \citet{ng2019facebook}.
We utilize beam search during decoding with the beam size 100, which is relatively large, since we want generated sentences to be more diverse while grammatically correct at the same time.
Note that the contrastive (BT) model is trained with the NT-Xent loss \cite{chen-etal-2020-a}, unlike CERT \cite{fang2020cert} which leverages the MoCo training objective \cite{he2020momentum}.

\subsection{SentEval Configurations} \label{subsec:senteval configurations}

\begin{table}[h]
\scriptsize
    \centering
    \begin{tabular}{l c}
    \toprule
    \textbf{Hyperparameters} & \textbf{Values} \\
    \midrule
    Random seed & 1, 2, 3, 4, 1234, 2345, 3456, 7890 \\
    K-fold & 10 \\
    Classifier (hidden dimension) & 50 \\ 
    Optimizer & Adam \\ 
    Batch size & 64 \\ 
    Tenacity & 5 \\
    Epoch & 4 \\ 
    \bottomrule
    \end{tabular}
    \caption{SentEval hyperparameters.}
    \label{table:table9}
\end{table}

In Table \ref{table:table9}, we stipulate the hyperparameters of the SentEval toolkit used in our experiment.
Additionally, we specify some minor modifications applied on our contrastive method (SG-OPT).
First, we use the portion of the concatenation of SNLI \cite{bowman2015large} and MNLI \cite{williams-etal-2018-broad} datasets as the training data instead of STS-B.
Second, we do not leverage the first several layers of PLMs when making positive samples, similar to \citet{wang2020sbert}, and utilize mean pooling instead of max pooling.

\subsection{GLUE Experiments} \label{subsec:glue experiments}

\begin{table}[h]
\scriptsize
    \centering
    \setlength{\tabcolsep}{0.15em}
    \begin{tabular}{l c c c c c}
    \toprule
    \textbf{Models} & \textbf{QNLI} & \textbf{SST2} & \textbf{COLA} & \textbf{MRPC} & \textbf{RTE} \\ 
    \midrule
    \textbf{BERT-base} & 90.97$_{\pm0.49}$ & 91.08$_{\pm0.73}$ & \textbf{56.63}$_{\pm3.82}$ & \textbf{87.09}$_{\pm1.87}$ & 62.50$_{\pm2.77}$ \\ 
    + \textbf{SG-OPT} & \textbf{91.28}$_{\pm0.28}$ & \textbf{91.68}$_{\pm0.41}$ & 56.36$_{\pm3.98}$ & 86.96$_{\pm1.11}$ & \textbf{62.75}$_{\pm3.91}$ \\
    \bottomrule
    \end{tabular}
    \caption{Experimental results on a portion of the GLUE validation set.}
    \label{table:table10}
\end{table}

We here investigate the impact of our method on typical supervised fine-tuning of BERT models.
Concretely, we compare the original BERT with one fine-tuned using our SG-OPT method on the GLUE \cite{wang2019glue} benchmark.
Note that we use the first 10\% of the GLUE validation set as the real validation set and the last 90\% as the test set, as the benchmark does not officially provide its test data.
We report experimental results tested on 5 sub-tasks in Table \ref{table:table10}.
The results show that our method brings performance improvements for 3 tasks (QNLI, SST2, and RTE).
However, it seems that SG-OPT does not influence much on supervised fine-tuning results, considering that the absolute performance gap between the two models is not significant. 
We leave more analysis on this part as future work.

\subsection{Representation Visualization (SBERT)} \label{subsec:representation visualization(SBERT)}

\begin{figure}[h]
    \begin{center}
    \includegraphics[width=0.9\linewidth]{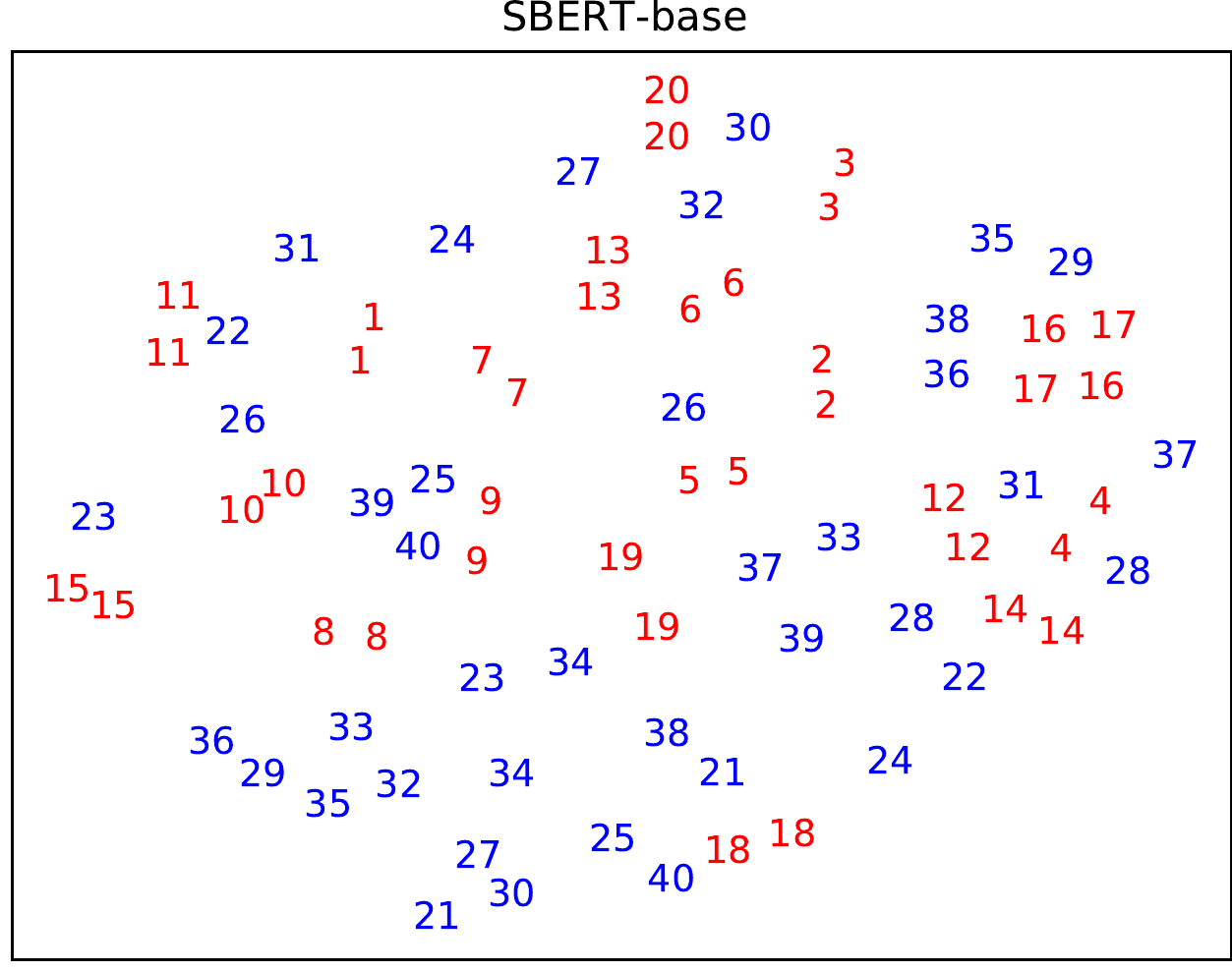}
    \caption{Visualization of sentence vectors computed by SBERT-base.} 
    \label{fig:figure6}
    \end{center}
\end{figure}

\begin{table*}[t!]
\scriptsize
    \centering
    \setlength{\tabcolsep}{0.4em}
    \begin{tabular}{l c c c c c c c c c}
    \toprule
    \textbf{Models} & \textbf{Pooling} & \textbf{STS}-B & \textbf{SICK-R} & \textbf{STS12} & \textbf{STS13} & \textbf{STS14} & \textbf{STS15} & \textbf{STS16} & \textbf{Avg.} \\ 
    \midrule
    \bf RoBERTa-base & & & & & & & & & \\
    + No tuning & CLS & 45.41 & 61.89 & 16.67 & 45.57 & 30.36 & 55.08 & 56.98 & 44.57 \\
    + No tuning & Mean & 54.53 & 62.03 & 32.11 & 56.33 & 45.22 & 61.34 & 61.98 & 53.36 \\
    + No tuning & WK & 35.75 & 54.69 & 20.31 & 36.51 & 32.41 & 48.12 & 46.32 & 39.16 \\
    + Contrastive (BT) & CLS & \textbf{79.93}$_{\pm1.08}$ & \textbf{71.97}$_{\pm1.00}$ & 62.34$_{\pm2.41}$ & 78.60$_{\pm1.74}$ & 68.65$_{\pm1.48}$ & 79.31$_{\pm0.65}$ & 77.49$_{\pm1.29}$ & \textbf{74.04}$_{\pm1.16}$ \\ 
    + \textbf{Contrastive (SG)} & CLS & 78.38$_{\pm0.43}$ & 69.74$_{\pm1.00}$ & \textbf{62.85}$_{\pm0.88}$ & 78.37$_{\pm1.55}$ & 68.28$_{\pm0.89}$ & \textbf{80.42}$_{\pm0.65}$ & \textbf{77.69}$_{\pm0.76}$ & 73.67$_{\pm0.62}$ \\
    + \textbf{Contrastive (SG-OPT)} & CLS & 77.60$_{\pm0.30}$ & 68.42$_{\pm0.71}$ & 62.57$_{\pm1.12}$ & \textbf{78.96}$_{\pm0.67}$ & \textbf{69.24}$_{\pm0.44}$ & 79.99$_{\pm0.44}$ & 77.17$_{\pm0.24}$ & 73.42$_{\pm0.31}$ \\
    \midrule
    \bf RoBERTa-large & & & & & & & & & \\
    + No tuning & CLS & 12.52 & 40.63 & 19.25 & 22.97 & 14.93 & 33.41 & 38.01 & 25.96 \\
    + No tuning & Mean & 47.07 & 58.38 & 33.63 & 57.22 & 45.67 & 63.00 & 61.18 & 52.31 \\
    + No tuning & WK & 30.29 & 28.25 & 23.17 & 30.92 & 23.36 & 40.07 & 43.32 & 31.34 \\
    + Contrastive (BT) & CLS & 77.05$_{\pm1.22}$ & 67.83$_{\pm1.34}$ & 57.60$_{\pm3.57}$ & 72.14$_{\pm1.16}$ & 62.25$_{\pm2.10}$ & 71.49$_{\pm3.24}$ & 71.75$_{\pm1.73}$ & 68.59$_{\pm1.53}$ \\ 
    + \textbf{Contrastive (SG)} & CLS & 76.15$_{\pm0.54}$ & 66.07$_{\pm0.82}$ & \textbf{64.77}$_{\pm2.52}$ & 71.96$_{\pm1.53}$ & 64.54$_{\pm1.04}$ & 78.06$_{\pm0.52}$ & 75.14$_{\pm0.94}$ & 70.95$_{\pm1.13}$ \\
    + \textbf{Contrastive (SG-OPT)} & CLS & \textbf{78.14}$_{\pm0.72}$ & \textbf{67.97}$_{\pm1.09}$ & 64.29$_{\pm1.54}$ & \textbf{76.36}$_{\pm1.47}$ & \textbf{68.48}$_{\pm1.58}$ & \textbf{80.10}$_{\pm1.05}$ & \textbf{76.60}$_{\pm0.98}$ & \textbf{73.13}$_{\pm1.20}$ \\
    \bottomrule
    \end{tabular}
    \caption{Performance of RoBERTa on STS tasks when combined with different sentence embedding methods. We could not report the performance of \citet{li-etal-2020-sentence} (Flow) as their official code do not support RoBERTa.}
    \label{table:table11}
\end{table*}

\subsection{RoBERTa's Performance on STS Tasks} \label{subsec:roberta's performance on sts tasks}

In Table \ref{table:table11}, we additionally report the performance of sentence embeddings extracted from RoBERTa using different methods.
Our methods, SG and SG-OPT, demonstrate their competitive performance overall.
Note that contrastive learning with back-translation (BT) also shows its remarkable performance in the case of RoBERTa-base.


\end{document}